%% file: main.tex
\documentclass[11pt]{article}
\usepackage{arxiv}

\usepackage{adjustbox}
\usepackage{wrapfig}
\usepackage{hyperref}       
\usepackage{url}            
\usepackage{booktabs}       
\usepackage{amsfonts}       
\usepackage{nicefrac}       
\usepackage{microtype}      
\usepackage{lipsum}
\usepackage{cite}
\usepackage{amsmath,amssymb,amsfonts}
\usepackage{algorithmic}
\usepackage{graphicx}
\usepackage{textcomp}
\usepackage{xcolor}
\usepackage{fancyhdr}
\usepackage[]{geometry}
\usepackage[]{ulem}
\usepackage{hyperref}
\usepackage{color}
\usepackage{soul}
\usepackage{tikz}
\usepackage{array}
\usepackage{fixltx2e}
\usepackage{stfloats}
\usepackage{verbatim}
\usepackage{subfiles}
\usepackage{booktabs}

\usepackage{caption}
\usepackage{subcaption}
\usepackage{floatrow}

\title{Data Engineering for Everyone}

\author{
  Vijay Janapa Reddi\\[5pt]
  Harvard University\\
   \And
   Greg Diamos \\[5pt]
   {Landing.AI} \\
   \And
   Pete Warden \\[5pt]
   {Google} \\
   \And
   Peter Mattson \\[5pt]
   {Google} \\
   \And
   David Kanter \\[5pt]
   {MLCommons} \\   [15pt]
}

\date{}

\begin{document}
\maketitle

\begin{abstract}
Data engineering is one of the fastest-growing fields within machine learning (ML). As ML becomes more common, the appetite for data grows more ravenous. But ML requires more data than individual teams of data engineers can readily produce, which presents a severe challenge to ML deployment at scale. Much like the software-engineering revolution, where mass adoption of open-source software replaced the closed, in-house development model for infrastructure code, there is a growing need to enable rapid development and open contribution to massive machine learning data sets. This article shows that open-source data sets are the rocket fuel for research and innovation at even some of the largest AI organizations. Our analysis of nearly 2000 research publications from Facebook, Google and Microsoft over the past five years shows the widespread use and adoption of open data sets. Open data sets that are easily accessible to the public are vital to accelerate ML innovation for everyone. But such open resources are scarce in the wild. So, can we accelerate data set creation and enable the rapid development of open data sets, akin to the rapid development of open-source software? Moreover, can we develop automatic data set generation frameowrks and tools to avert the data scarcity crisis?
\end{abstract}

\input{intro}

\input{opendataset}
\input{dataengineering}
\input{calltoaction}\newpage
\input{bio}

\input{acknowledgements}

\newpage
\bibliographystyle{ieeetr}
\bibliography{references}

\input{appendix}

\end{document}

%% file: intro.tex
\section{Introduction}

The rise of open-source software necessitated a software-engineering revolution (new standards, tools, licenses, etc.) to overcome the problems facing large distributed teams working on enormous code bases. Today, machine learning (ML) builds atop this vibrant and creative open-source ecosystem and toolchain - leading frameworks such as TensorFlow, PyTorch and Keras are open and boast thousands of contributors around the globe. But ML is utterly unlike explicitly written software and relies heavily on data to implicitly define program behavior. Shifting the importance from explicit code to data introduces an even greater and more challenging problem: too little data.

A whole new set of engineering challenges is emerging as massive data sets become vital to creating ML applications. We have yet to figure out all the tools, practices and patterns to effectively handle billions of training examples in a large system. We believe it will require a whole new discipline: \textit{data engineering}. Data engineering is not new; it's a fast-growing IT area, boasting 50\% year-over-year job growth thanks partly to the insatiable need for data to train ML models~\cite{datanami_2020}. 

A typical data engineer aids in creating data sets to train high-quality models. And let's not confuse data scientists with data engineers. Data scientists handle data analytics—that is, the science of transforming the processed data into information. Data engineers are behind the scenes architecting, building, managing and operating the data-creation pipelines that set the stage for data scientists. As ML becomes more common, the appetite for data will grow more ravenous. It requires more data than data engineers can readily produce, presenting a challenge for ML deployment. 

Although Linux can operate by itself, for example, an image classifier such as ResNet is a mere curiosity unless it incorporates training data. ML solutions must combine software and data, with data fueling the software. The ML community does build data sets, but this endeavor is often artisanal—painstaking and expensive. We lack the high productivity and efficient tools to make building data sets easier, cheaper and more repeatable. Constructing a data set today is akin to programming before the era of compilers, libraries and other tools. So, the question is whether we can accelerate data-set creation by democratizing data engineering—that is, building computer systems (“architectures”) that can automatically produce training data cheaply and efficiently.

%% file: opendataset.tex
\section{Open Data Sets}
\label{sec:open_data_sets}
Demand for data engineering is soaring, but data sets are often the domain of large organizations with billions of users owing to their proprietary nature, as well as licensing and privacy issues. These data sets are often inaccessible to individual developers, researchers and students, impeding progress, and this leads us to ask a fundamental question. \textit{What if we could enable the rapid development of open data sets, akin to how the community has enabled the rapid development of open-source software? }

Some influential data sets are open. For example, ImageNet\cite{imagenet} (first released in 2009) fueled image classification with its corpus of 14 million images. MS-COCO\cite{coco} (first released in 2014) provides 1.5 million object instances to enable object-detection research. And SQuAD\cite{squad} (first released in 2016) helped stimulate natural-language-processing (NLP) research by offering a reading-comprehension data set containing more than 100,000 questions. These examples helped spark the ML revolution, which led to a hardware and software renaissance that unlocked amazing new applications. 

To quantify the impact of open-source data sets, we looked at ML-research publications from some leading AI organizations: Facebook AI Research (FAIR)\cite{facebook}, Google AI\cite{google} and Microsoft AI\cite{microsoft}. We looked in relevant areas, such as computer vision and NLP. Our consideration included major venues, such as ICLR, ICML, NeurIPS and CVPR conferences. If a paper used a data set, we conducted a search to determine whether the data set is open access (e.g., whether it has a GitHub repository or a dedicated download web page). If so, we counted it as using an open data set.

\begin{figure}[t]
    \includegraphics[width=.6\textwidth]{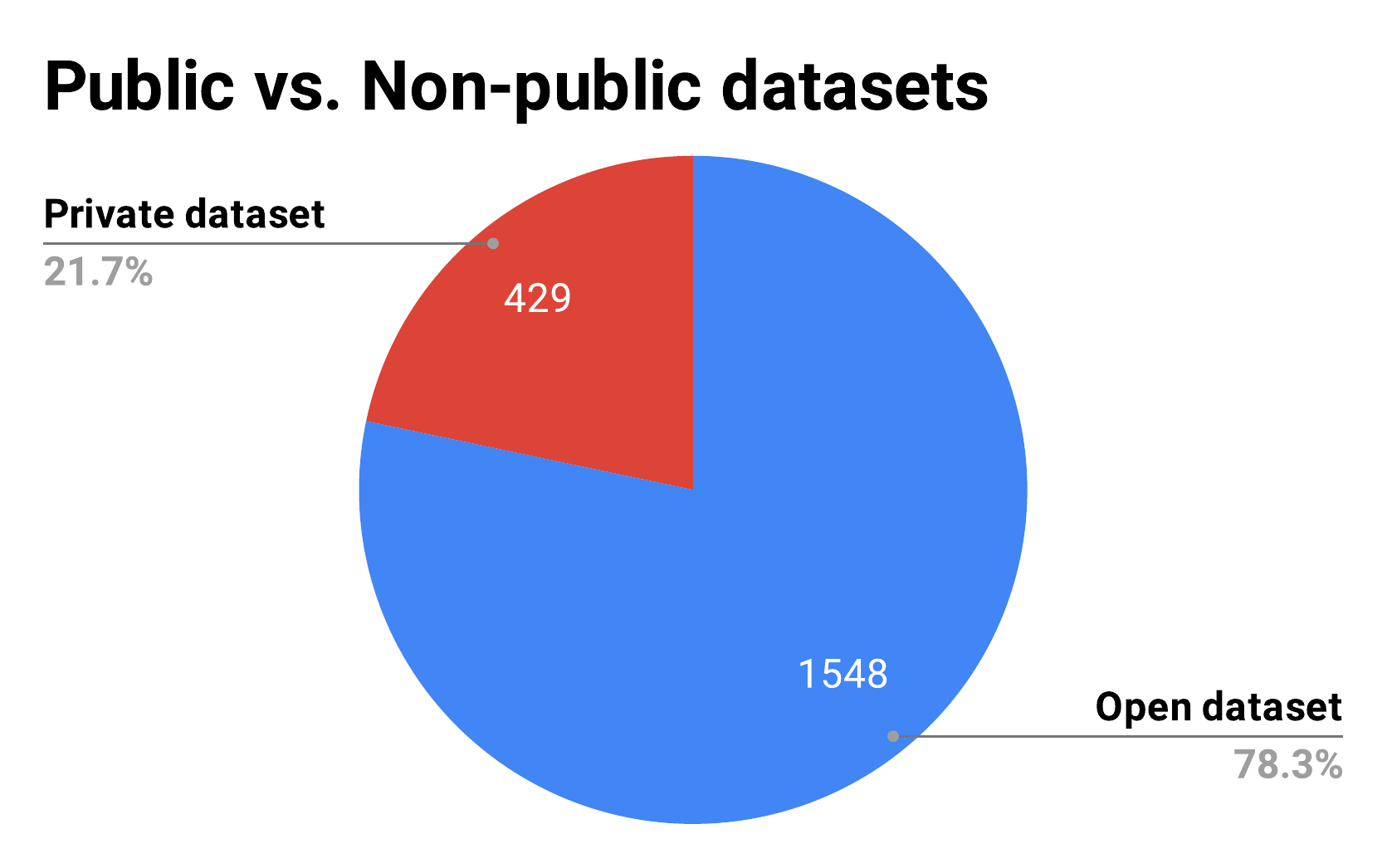}
    \caption{Percentage of ML research papers that rely on open versus nonpublic data sets.}
    \label{fig:percentage_ML}
\end{figure}

Our analysis presently includes nearly 2,000 papers from the three above mentioned organizations, published between 2015 and the start of 2021 (i.e., post-AlexNet). Figure~\ref{fig:percentage_ML} summarizes our main findings. We determined that a staggering 78.3\% of industry research publications employ open data sets to report their findings. Only 21.7\% use internal proprietary data sets for reporting.

\begin{figure}[t]
    \begin{subfigure}{.48\linewidth}
        \includegraphics[width=\textwidth]{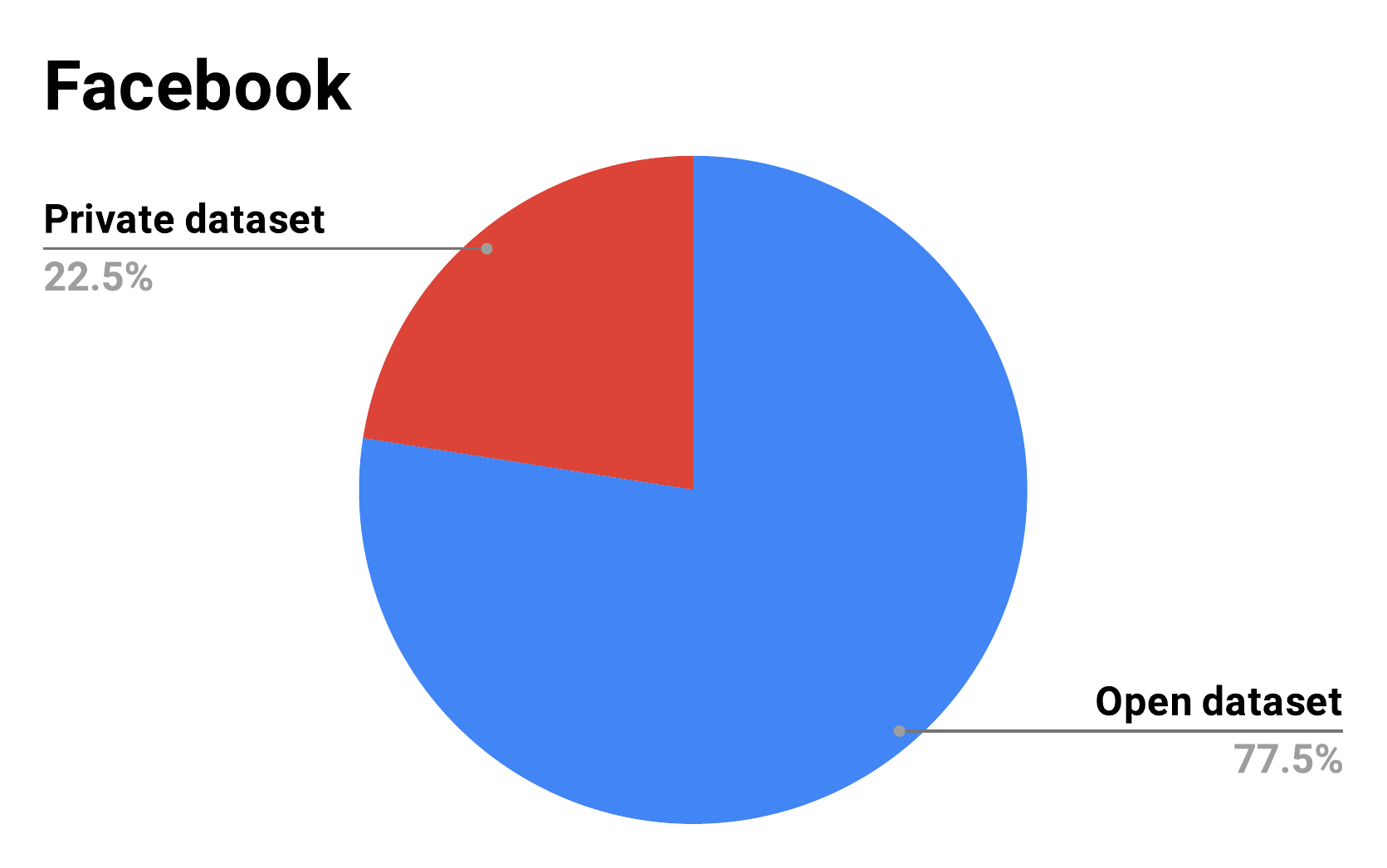}
    \end{subfigure}
    \begin{subfigure}{.48\linewidth}
        \includegraphics[width=\textwidth]{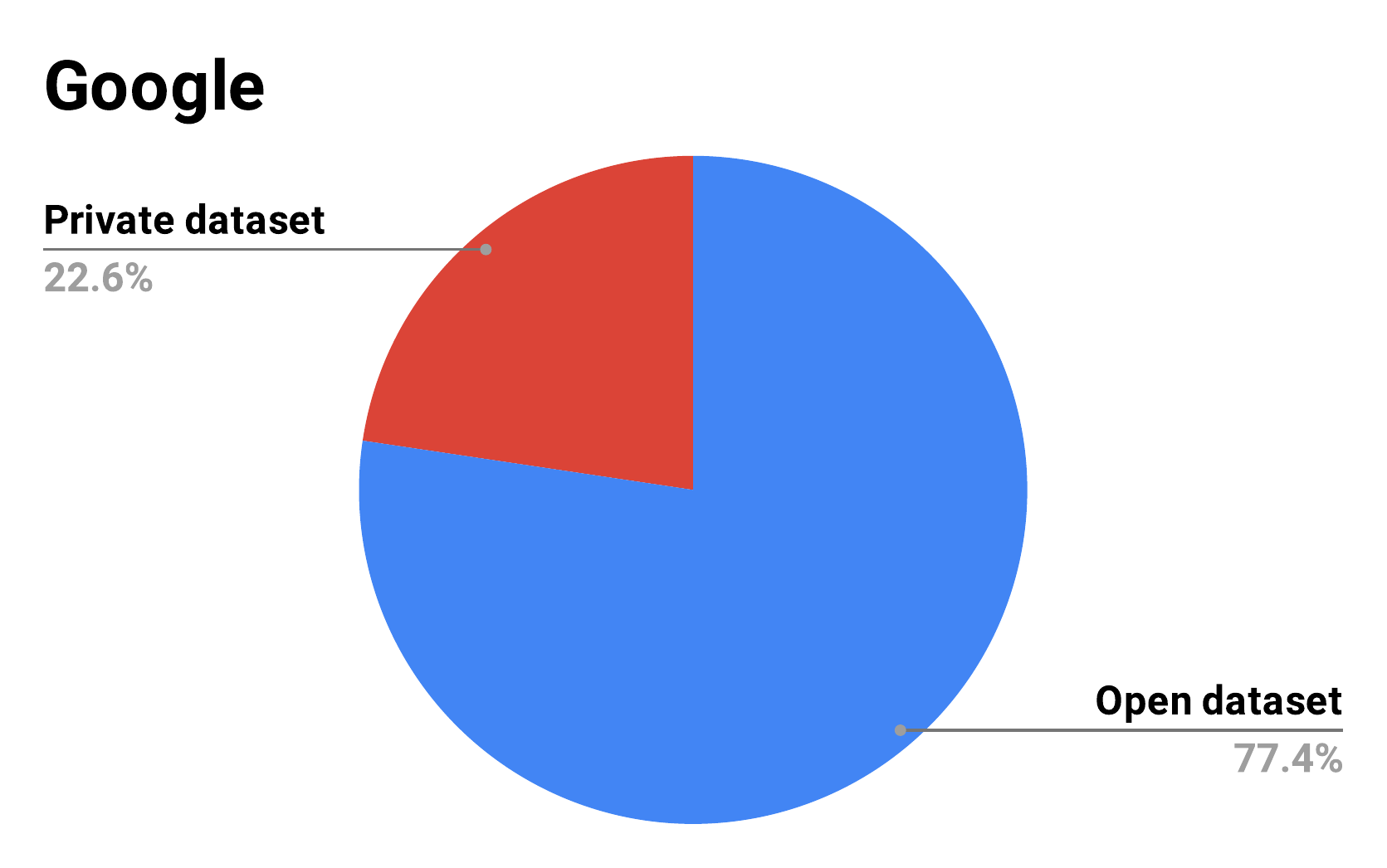}
    \end{subfigure}
    \begin{subfigure}{.48\linewidth}
        \includegraphics[width=\textwidth]{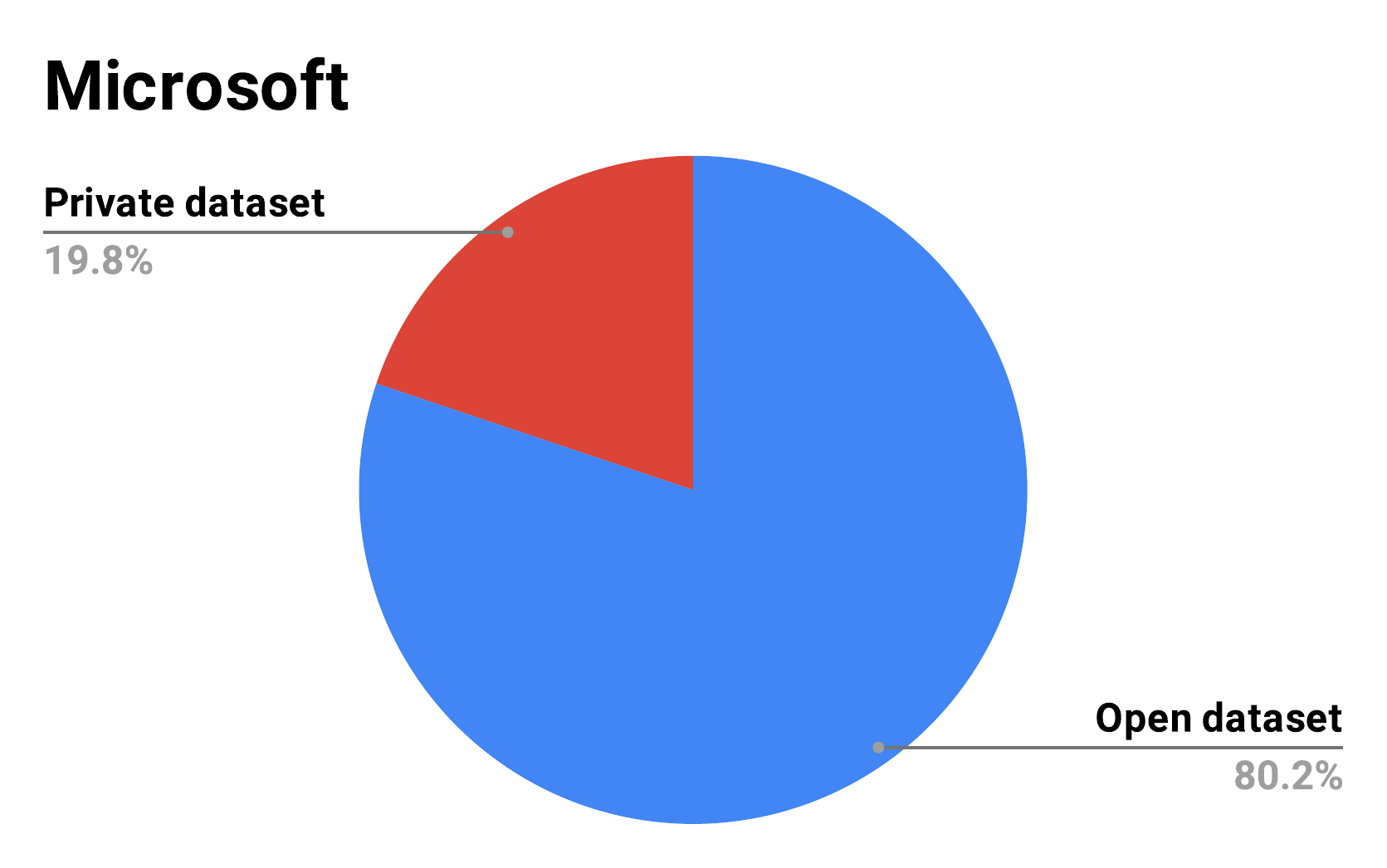}
    \end{subfigure}
    \caption{Breakdown the percentage of ML research papers that rely on open versus nonpublic data sets across various AI organizations. The use of open data sets is pervasive.}
    \label{fig:percentage_ML_breakdown}
\end{figure}

We further analyzed the breakdown of the data by the individual organizations. Figure~\ref{fig:percentage_ML_breakdown} shows the organizational level breakdown. Our findings remain consistent. All three AI vendors rely extensively on open data sets for demonstrating research advancements. Interestingly, the use of open data sets is increasing. As ML research accelerates, more industry publications are joining in. 

Leading researchers use open data sets for various reasons. The most obvious is simply accessibility. Open data sets are publicly hosted, are free to use and have a broad audience—the gold standard for reproducibility and verifiability. Bootstrapping ML research therefore becomes quick and easy. 

Open data sets provide a common platform for systematically assessing ML innovation. Open data sets bring about the values around transparency, innovation, democratization and efficiency improvements. As mentioned above, transparency aids with fair and useful results reproduction. ML models thrive on data sets, and as such they nurture innovation; open data sets enable the community as a whole to thrive. Ultimately, they lead to efficiency improvements because data sets lead to improved models that push hardware capabilities' efficiency limits. We note the value of putting in the extra effort to add the additional metadata to open datasets to make them easily searchable and accessible, as described by the FAIR principles for scientific data management~\cite{wilkinson2016fair}.

Given these obvious benefits, the next big question is whether the data-engineering practices we have followed so far are sufficient to fuel future creation of open-source data sets at the rate needed.

\begin{figure}[t]
    \begin{subfigure}{.49\linewidth}
        \includegraphics[width=\textwidth]{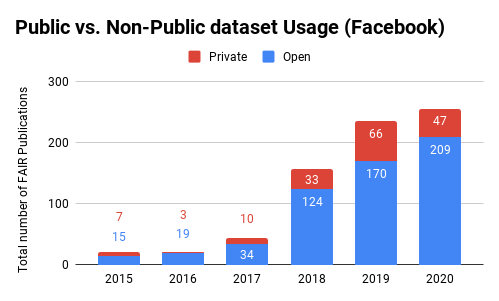}
    \end{subfigure}
    \hfill
    \begin{subfigure}{.49\linewidth}
        \includegraphics[width=\textwidth]{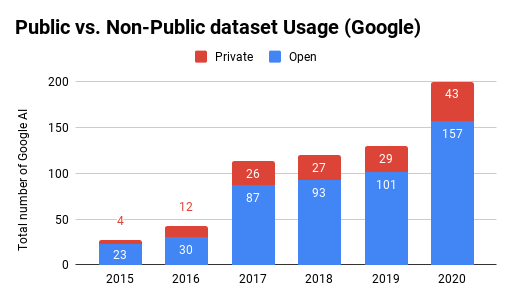}
    \end{subfigure}
    \begin{subfigure}{.49\linewidth}
        \includegraphics[width=\textwidth]{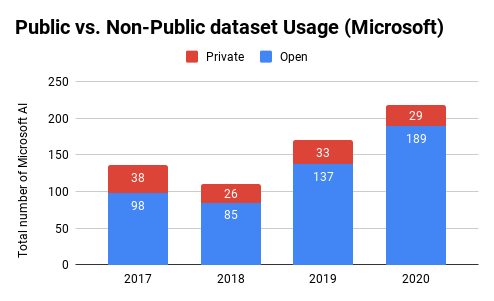}
    \end{subfigure}
    \caption{Ratio of ML research papers that rely on public versus private data sets per year. As the research output goes up, so does the usage of the open source data sets.}
    \label{fig:ratio_ML}
\end{figure}

%% file: dataengineering.tex
\section{Data Engineering}
\label{sec:data_engineering}
People have been working with large ML data sets for decades, but in many cases that work has been a sideline for exploring architectural or algorithmic problems. Data engineering produces repeatable and scalable techniques that can be shared and yield superior results quicker and more cost-effectively than artisanal approaches. Only in the last few years has data-set creation employed this kind of systematic approach. There is a growing need for more of such approaches.

The development of ImageNet\cite{imagenet}, MovieLens\cite{movielens}, DeepSpeech\cite{deepspeech}, CheXpert\cite{chexpert} and other well-known data sets is revealing; an especially illustrative example is ImageNet, developed by students who lacked the resources to solve a seemingly insurmountable problem. Around 2005, researchers at Princeton attempted to manually find 100 images for each of the 1.5 million words in WordNet. But they estimated the project would take practically forever. Over the next several years, a few computer-vision researchers considered how they could perform the task more efficiently. 

Using exclusively open-source technologies such as the Internet, information retrieval, computer-aided design, and computer vision, these researchers labeled about 10 million images for ImageNet on a budget of less than \$40,000. In doing so, they helped kick off the deep-learning revolution for image classification tasks, which paved a significant way for AI’s widespread adoption.

Despite this fantastic success, we have forgotten the lesson of these data-engineering pioneers. Small groups produced high-value data sets using only 2005-era workstations. They did so by employing technologies that dramatically increased their productivity. Today, despite vast software infrastructure (most of it open source) for developing models like AlexNet, practically nothing is available for building open data sets like ImageNet. When running on Nvidia’s A100 or another optimized AI chip, deep-learning frameworks such as TensorFlow and PyTorch make the old AlexNet look ancient. But the technologies that the ImageNet creators employed are still bleeding edge compared with the resource-intensive human labeling pipelines that frequently serve in the industry. ImageNet used Internet-search technology to expand WordNet into a custom web crawler. One student could reason about the space of interesting images and locate them across the Internet with a single query.  Specially designed processes could label hundreds of images at once for just a few cents. Semi-supervised learning algorithms focused the limited labeling capacity on the most promising areas. So this begs the question---how can we accelerate the pace of data set creation?

%% file: calltoaction.tex
\section{A Call to Action}
\label{sec:call_to_action}
The time has come for the community to start adopting open-source strategies to enable the rapid development of open data sets. It is time to start documenting the real-world process of data-set creation, specifically so that we can begin the process of spotting patterns and curating techniques and building systems that efficiently generate data sets. This task is especially hard for academic researchers, who rely on others to create and curate the much-needed open data sets. 

Efforts to overcome that difficulty have included identifying areas for which improved training data can have a significant impact; speech is one example. MLCommons\cite{mlcommons}, a community-driven ML-systems community, has therefore recently begun working on the ``\textit{People’s Speech}''\cite{mlcommonspeople} data set, which is 100x bigger than any other available publicly. It includes over 87,000 hours of transcribed speech. Ample public-domain speech data is on the Internet, but the task of labeling it manually has a projected price tag of \$10 million. The data-engineering question we set out to answer is whether we can democratize open data-set generation through computer-system innovation. Can we economically and responsibly collect and maintain useful data sets for future AI systems?

Drawing inspiration from ImageNet, we’re using emerging technology to make our labelers more productive by leveraging existing compute and ML tools. As one example, we use a generative text model to create a novel text corpus that captures a very large vocabulary.  We then vocalize this text data using an ML-based speech synthesis system to construct explicitly labeled speech examples. Using this approach and a variety of other techniques,  we have collected and labeled around 87,000 hours of new data on a budget of less than \$20,000.

At the other end of the spectrum is the ``\textit{1000 Words in 1000 Languages}''\cite{speechcommands} project. The 1000 Words in 1000 Languages project involves automating the construction of custom speech-command data sets (e.g., spoken keywords such as  ``Alexa'') for ML-enabled IoT devices. Using open data sets such as Common Voice\cite{voice} and off-the-shelf \textit{forced alignment} algorithms \cite{montreal}, we’re building a data-engineering pipeline that can take any recorded speech and automatically generate per-word audio clips to train keyword-spotting models. Usually, humans generate these clips manually—a costly and time-consuming task. Instead, we use the estimated timings from forced alignment to automatically extract words, greatly expanding the number of words we cover. 

We are still in the early stages of automatic open data set generation. There are a wide array of challenges that remain. Some involve defining the metrics for assessing the data sets that automatically emerge from these pipelines. Example metrics include licensing, size, quality and ethical considerations, and qualification. Industrial users, for instance, require licenses for commercial use. The rise of common licenses such as CC-BY and CC-0 has helped, similar to how BSD and Apache2 have aided open-source software. But efficiently locating and verifying licenses for millions of training-data samples remains problematic. Another emerging area is ethics. For example, even if a picture is freely available on the Internet under CC-0 or another permissive license, its owner may object to its use in improving face-identification systems for mass surveillance.

Five years from now, imagine a world where we assess not how well a system performs AI tasks but how well the computer system performs at processing raw data to generate practical, verifiable data sets for training ML models. Our People’s Speech and Speech Commands projects are both community-driven efforts to increase data-engineering efficiency. We hope they’ll be arenas for testing and improving novel data-engineering systems and ideas. 

We welcome the community's input and assistance in democratizing data engineering. Students, researchers, enthusiasts and organizations are all welcome. Academia can help directly by involvement of faculty and students in our projects but also broadly by emphasizing data engineering in the curriculum. Programs in data engineering are less prevalent than those in data science and many students do not realize that there are a large number of jobs in this area. Join our effort at MLCommons. Additional details can be found here: \url{https://mlcommons.org/en/peoples-speech/}. 

MLCommons aims to accelerate machine learning innovation to benefit everyone. Machine learning has tremendous potential to save lives in areas like healthcare and automotive safety and to improve information access and understanding through technologies like voice interfaces, automatic translation, and natural language processing. However, machine learning is entirely unlike conventional software -- developers train an application rather than program it -- and requires a whole new set of techniques analogous to the breakthroughs in precision measurement, raw materials, and manufacturing that drove the industrial revolution. MLCommons aims to answer the needs of the nascent machine learning industry through open, collaborative engineering in three areas: benchmarks, datasets and best practices.

%% file: bio.tex
\section*{Author Biographies}
\label{sec:bio}

\begin{table}[h]
\begin{tabular}{p{.2\linewidth}p{.7\linewidth}}

\vtop{%
  \vskip0pt
  \hbox{%
  \includegraphics[height=1in]{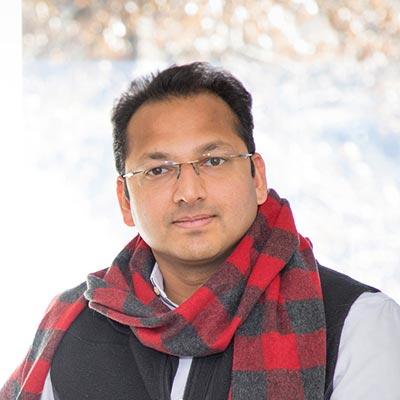}%
}
}
 &  \vspace{0pt}Vijay Janapa Reddi is an Associate Professor at Harvard University, Inference Co-chair for MLPerf, and a founding member of MLCommons, a nonprofit ML organization that aims to accelerate ML innovation. He also serves on the MLCommons board of directors.
\\
\vtop{%
  \vskip-2ex
  \hbox{%
  \includegraphics[height=1in]{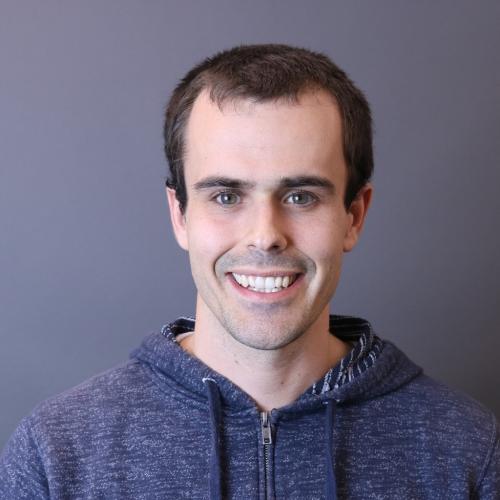}
}%
}
 &  Greg Diamos leads transformation engineering at Landing AI. He’s a data-set chair and founding member of MLCommons.
\\
\vtop{%
  \vskip-2ex
  \hbox{%
  \includegraphics[height=1in]{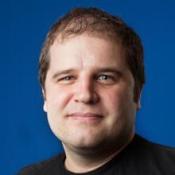}%
}%
}
 &  Pete Warden is Technical Lead of TensorFlow Mobile and Embedded at Google.
\\
\vtop{%
  \vskip-2ex
  \hbox{%
  \includegraphics[height=1in]{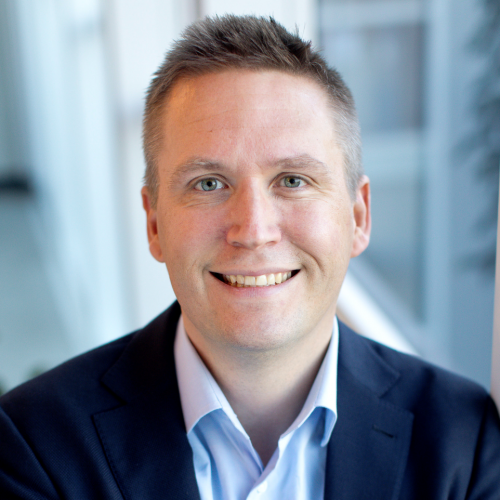}
}%
}
 &  Peter Mattson leads ML Metrics at Google. He is the General Chair and co-founder of MLPerf, and President and founding member of MLCommons.
\\
\vtop{%
  \vskip-2ex
  \hbox{%
  \includegraphics[height=1in]{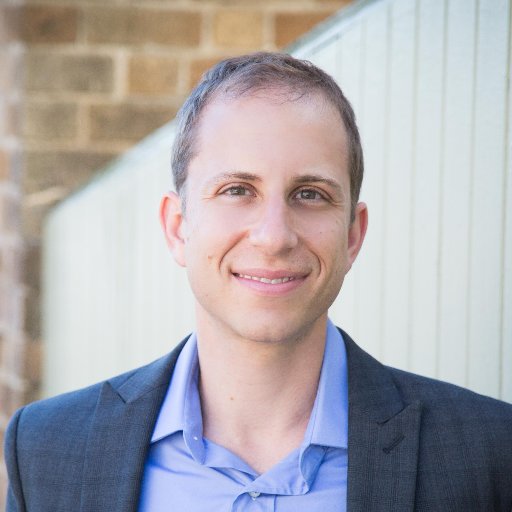}
}%
}
 & David Kanter is Executive Director, a founding member of MLCommons, and previously served as Power chair and Inference co-chair for MLPerf.
\\
\end{tabular}
\end{table}

%% file: acknowledgements.tex
\section*{Acknowledgements}
\label{sec:acknowledgement}
We are grateful to Zishen Wan from Harvard University/Georgia Tech for collecting the data that was assimilated into the graphs presented in this report.

%% file: appendix.tex
\renewcommand{\thesubsection}{\Alph{subsection}}
\newpage
\section*{Appendix}
\label{sec:appendix}
Here, we list several details about our public/non-public dataset collection method and raw data that shown in Section~\ref{sec:open_data_sets}. To the best of our ability, we provide sufficient details to reproduce our results and address common clarification questions.

\textbf{Dataset collection range:} 
Published papers across three corporations (Facebook, Google, Microsoft) and recent five years (2015-2020).

\textbf{Dataset collection methods and results:} 
\begin{itemize}
    \item General method: 
    \begin{itemize}
        \item  Step 1: Within one specific paper, we search “dataset”, “data set” or related terms to check whether and what datasets are used in this paper. 
        \item Step 2: If there is a dataset used in the paper, we conduct a Internet search to determine whether the dataset is open-sourced, e.g., have a github repo or have a dedicated download webpage, etc. If so, we count this paper using public datasets, otherwise count it as not using public dataset.
    \end{itemize}
    \item Facebook:
    \begin{itemize}
        \item Paper sources: \url{https://research.fb.com/publications/}
        \item Method: Count papers in \textit{Computer Vision}, \textit{Machine Learning}, and \textit{Natural Language Processing \& Speech} categories.
        \item Results: we look into 737 papers across six years in total, the count of paper using public/non-public dataset is listed in Table~\ref{tab:facebook}.
        
\begin{table}[h]
\centering
\resizebox{0.6\columnwidth}{!}{%
\begin{tabular}{|c|c|c|c|}
\hline
\textbf{Year} & \textbf{Using Public Dataset} & \textbf{Not Using Public Dataset} & \textbf{Public Ratio}  \\ \hline
\textit{2015} & 15 & 7 & 68.18\%  \\ 
\textit{2016} & 19 & 3 & 86.36\%  \\ 
\textit{2017} & 34 & 10 & 77.27\%  \\ 
\textit{2018} & 124 & 33 & 78.98\%  \\ 
\textit{2019} & 170 & 66 & 72.03\%  \\
\textit{2020} & 209 & 47 & 81.64\%  \\\hline
\textit{Total} & 571 & 166 & 77.48\% \\ \hline
\end{tabular}%
}
\caption{The count of publications using public/non-public dataset (Facebook).}
\label{tab:facebook}
\end{table}
    \end{itemize}
    \item Google:
    \begin{itemize}
        \item Paper sources: \url{https://research.google/pubs/}
        \item Method: Count papers published in \textit{ICLR}, \textit{ICML}, \textit{NeurIPS}, and \textit{CVPR} conferences.
        \item Results: we look into 605 papers across five years in total, the count of paper using public/non-public dataset is listed in Table~\ref{tab:google}.
        
        \begin{table}[h]
\centering
\resizebox{0.6\columnwidth}{!}{%
\begin{tabular}{|c|c|c|c|}
\hline
\textbf{Year} & \textbf{Using Public Dataset} & \textbf{Not Using Public Dataset} & \textbf{Public Ratio}  \\ \hline
\textit{2015} & 23 & 4 & 85.19\%  \\ 
\textit{2016} & 30 & 12 & 71.43\%  \\ 
\textit{2017} & 87 & 26 & 76.99\%  \\ 
\textit{2018} & 93 & 27 & 77.50\%  \\ 
\textit{2019} & 101 & 29 & 77.69\%  \\
\textit{2020} & 157 & 43 & 78.50\%  \\\hline
\textit{Total} & 468 & 137 & 77.36\% \\ \hline
\end{tabular}%
}
\caption{The count of publications using public/non-public dataset (Google).}
\label{tab:google}
\end{table}
        
      \end{itemize}   

    \item Microsoft:
    \begin{itemize}
        \item Paper sources: \url{https://www.microsoft.com/en-us/research/}
        \item Method: Count papers in \textit{Computer Vision} and \textit{Human Language Technology} categories.
        \item Results: we look into 635 papers across three years in total, the count of paper using public/non-public dataset is listed in Table~\ref{tab:microsoft}.
        
        \begin{table}[h]
\centering
\resizebox{0.6\columnwidth}{!}{%
\begin{tabular}{|c|c|c|c|}
\hline
\textbf{Year} & \textbf{Using Public Dataset} & \textbf{Not Using Public Dataset} & \textbf{Public Ratio}  \\ \hline
\textit{2017} & 98 & 38 & 72.06\%  \\ 
\textit{2018} & 85 & 26 & 76.58\%  \\ 
\textit{2019} & 137 & 33 & 80.59\%  \\
\textit{2020} & 189 & 29 & 86.70\%  \\\hline
\textit{Total} & 509 & 126 & 80.16\% \\ \hline
\end{tabular}%
}
\caption{The count of publications using public/non-public dataset (Microsoft).}
\label{tab:microsoft}
\end{table}
        
      \end{itemize}

\end{itemize}